%% file: main.tex
\RequirePackage[svgnames]{xcolor}

\documentclass[11pt,letterpaper]{mystyle}

\usepackage[all]{hypcap}
\usepackage[comma,authoryear,compress]{natbib}
\usepackage{hyperref}
\hypersetup{
    colorlinks = true,
    citecolor = {YaleBlue},
}

\usepackage{algorithm}
\usepackage{algorithmicx}
\usepackage{algpseudocode}
\usepackage{microtype}
\usepackage{graphicx}
\expandafter\def\csname ver@subfig.sty\endcsname{}
\usepackage{booktabs}
\usepackage{float}
\usepackage{bigstrut}

\usepackage{amsmath}
\usepackage{amssymb}
\usepackage{mathtools}
\usepackage{amsthm}
\usepackage{mathrsfs}
\usepackage{nicefrac}
\usepackage{dsfont}
\usepackage{enumitem}
\usepackage{subcaption}
\usepackage{cleveref}

\usepackage{siunitx}
\usepackage{tabularx}
\usepackage{colortbl}
\usepackage{makecell}
\usepackage{array}

\usepackage[utf8]{inputenc}
\usepackage[T1]{fontenc}
\usepackage{url}
\usepackage{amsfonts}
\usepackage{wrapfig}
\usepackage{stackengine}
\usepackage[font=small,labelfont=bf]{caption}
\usepackage{adjustbox}
\usepackage{rotating}
\usepackage{pifont}

\input{macro}

\newtcolorbox{AIbox}[2][]{aibox,title=#2,#1}
\definecolor{lightblue}{rgb}{0.22,0.45,0.70}
\definecolor{Gray}{gray}{0.95}
\definecolor{Cornsilk}{rgb}{1.0, 0.97, 0.86}
\definecolor{darkgreen}{RGB}{0, 150, 0}

\newcommand{\cmark}{\textcolor{darkgreen}{\ding{51}}}
\newcommand{\xmark}{\textcolor{red}{\ding{55}}}

\title{VLA-OPD: Bridging Offline SFT and Online RL for Vision-Language-Action Models via On-Policy Distillation}

\runningtitle{VLA-OPD: Bridging Offline SFT and Online RL for Vision-Language-Action Models via On-Policy Distillation}

\author{
  Zhide Zhong$^1$,
  Haodong Yan$^1$,
  Junfeng Li$^1$,
  Junjie He$^1$,
  Tianran Zhang$^1$, and
  Haoang Li
}

\affil[1]{HKUST (GZ)}

\begin{document}

\input{sections/abstract}

\maketitle
\vspace{3mm}
\input{sections/introduction}
\input{sections/preliminaries}
\input{sections/method}
\input{sections/experiments}
\input{sections/relatedwork}
\input{sections/conclusion}
\clearpage
\bibliography{main}

\end{document}

%% file: macro.tex
\definecolor{blanchedalmond}{rgb}{1.0, 0.92, 0.8}
\definecolor{carmine}{rgb}{0.59, 0.0, 0.09}
\definecolor{lightblue}{rgb}{0.22,0.45,0.70}%

\renewcommand{\mathbf}{\boldsymbol}

\makeatletter
\def\Ddots{\mathinner{\mkern1mu\raise\p@
\vbox{\kern7\p@\hbox{.}}\mkern2mu
\raise4\p@\hbox{.}\mkern2mu\raise7\p@\hbox{.}\mkern1mu}}
\makeatother

\definecolor{amaranth}{rgb}{0.9, 0.17, 0.31}
\definecolor{antiquebrass}{rgb}{0.8, 0.58, 0.46}
\definecolor{antiquefuchsia}{rgb}{0.57, 0.36, 0.51}
\definecolor{chromeyellow}{rgb}{0.31, 0.47, 0.26}

%% file: sections/abstract.tex
\begin{abstract}
Although pre-trained Vision-Language-Action (VLA) models exhibit impressive generalization in robotic manipulation, post-training remains crucial to ensure reliable performance during deployment. However, standard offline Supervised Fine-Tuning (SFT) suffers from distribution shifts and catastrophic forgetting of pre-trained capabilities, while online Reinforcement Learning (RL) struggles with sparse rewards and poor sample efficiency. In this paper, we propose On-Policy VLA Distillation (VLA-OPD), a framework bridging the efficiency of SFT with the robustness of RL. Instead of relying on sparse environmental rewards, VLA-OPD leverages an expert teacher to provide dense, token-level supervision on the student’s self-generated trajectories. This enables active error correction on policy-induced states while preserving pre-trained general capabilities through gentle alignment. Crucially, we formulate VLA-OPD via a Reverse-KL objective. Unlike standard Forward-KL that induces mode-covering entropy explosion, or Hard-CE that causes premature entropy collapse, our bounded mode-seeking objective ensures stable policy learning by filtering out the teacher's epistemic uncertainty while maintaining action diversity. Experiments on LIBERO and RoboTwin2.0 benchmarks demonstrate that VLA-OPD significantly improves sample efficiency over RL and robustness over SFT, while effectively mitigating catastrophic forgetting during post-training.
\vspace{2mm}


\vspace{1mm}

\textit{Project Page: \url{https://irpn-lab.github.io/VLA-OPD/}}

\vspace{2mm}

\end{abstract}

%% file: sections/introduction.tex
\vspace{-4mm}
\section{Introduction}
\label{sec:intro}

The integration of Large Language Models (LLMs) with visual perception has catalyzed a paradigm shift in embodied intelligence, giving rise to generalist Vision-Language-Action (VLA) models~\cite{kim2024openvla,zitkovich2023rt,bai2025towards,intelligence2025pi06vlalearnsexperience,intelligence2025pi05visionlanguageactionmodelopenworld,gr00tn1_2025}. By unifying perception, planning, and control into a single transformer architecture, pre-trained VLAs exhibit remarkable generalization across diverse environments and instructions. However, despite their broad capabilities, directly deploying pre-trained foundation models often struggles with precise execution in specific downstream tasks. Consequently, to translate this generalist knowledge into reliable, deployable robotic policies, post-training has emerged as an essential step in adapting VLA models.

Currently, the landscape of VLA post-training is dominated by two primary paradigms: offline Supervised Fine-Tuning (SFT) and online Reinforcement Learning (RL). SFT, typically implemented as behavior cloning, maximizes the likelihood of expert actions given static observation histories~\cite{o2024open,black2024pi0visionlanguageactionflowmodel}. With dense supervision at every token, SFT is optimization-stable and fast to converge; however, it inherently demands large-scale, high-quality expert demonstrations to cover diverse scene distributions. To reduce this reliance on static datasets, recent efforts have explored online RL for VLA post-training~\cite{li2025simplevla,zang2025rlinf,lu2025vla,liu2025can,xu2025stare,xiao2025self}. By allowing the model to interact with the environment and optimizing for task success, RL exposes the policy to its own induced state distribution, aiming to improve closed-loop robustness through active exploration and error correction.

\begin{table}[t]
\footnotesize
    \centering
    \caption{\textbf{Comparison of VLA training paradigms.} VLA-OPD combines the best of both worlds: it inherits the \textbf{Few-Demo capability} of RL (learning from limited expert trajectories) while maintaining the \textbf{Fast Convergence} of SFT (via dense supervision).}
    \label{tab:comparison}
    \setlength{\tabcolsep}{3.5pt} 
    \resizebox{0.95\linewidth}{!}{
    \begin{tabular}{lcccccc} 
        \toprule
        \textbf{Paradigm} & \textbf{Sampling} & \textbf{Signal} & \textbf{Few-Demo} & \textbf{Convergence} & \textbf{Anti-forget} & \textbf{Robustness} \\
        \midrule
        VLA-Offline SFT & Off-policy & \textcolor{darkgreen}{Dense} & \textcolor{red}{\xmark} & \textcolor{darkgreen}{Fast} & \textcolor{red}{\xmark} & \textcolor{red}{\xmark} \\
        VLA-Online RL & \textcolor{darkgreen}{On-policy} & Sparse & \textcolor{darkgreen}{\cmark} & \textcolor{red}{Slow} & \textcolor{darkgreen}{\cmark} & \textcolor{darkgreen}{\cmark} \\
        \midrule
        \textbf{VLA-OPD (Ours)} & \textcolor{darkgreen}{\textbf{On-policy}} & \textcolor{darkgreen}{\textbf{Dense}} & \textcolor{darkgreen}{\cmark} & \textcolor{darkgreen}{\textbf{Fast}} & \textcolor{darkgreen}{\cmark} & \textcolor{darkgreen}{\cmark} \\
        \bottomrule
    \end{tabular}
    }
    \vspace{-0.5cm}
\end{table}

Despite their respective strengths, both paradigms suffer from critical limitations. SFT is fundamentally constrained by its ``off-policy'' nature; it is highly vulnerable to \textbf{distribution shifts} and often suffers from \textbf{catastrophic forgetting} due to aggressive parameter updates on static, disjoint datasets~\cite{zhu2025path,lai2025reinforcement,chu2025sft,shenfeld2025rl}. Conversely, \textbf{online RL} addresses distribution shifts via environment interaction but relies on sparse rewards, resulting in prohibitive sample inefficiency and high-variance optimization~\cite{li2025simplevla}. Furthermore, simply adapting SFT to an on-policy setting (e.g., DAGGER~\cite{kelly2019hg}) typically relies on suboptimal alignment objectives. Using a Forward-KL divergence forces a \textbf{mode-covering} behavior that mimics the teacher's epistemic uncertainty, leading to an \textbf{entropy explosion}. Conversely, employing Hard-CE (argmax matching) causes \textbf{premature entropy collapse}, depriving the student of the action diversity needed for effective state-space exploration.

To bridge this gap, we introduce \textbf{On-Policy VLA Distillation (VLA-OPD)}, a unified framework that synthesizes the efficiency of SFT with the robustness of RL (summarized in Table~\ref{tab:comparison}). VLA-OPD leverages an expert teacher to provide dense, token-level supervision on the student's self-generated trajectories, intrinsically enabling active error recovery without sparse rewards. Crucially, we formulate VLA-OPD using a \textbf{Reverse-KL} objective to overcome the aforementioned optimization flaws. By promoting bounded \textbf{mode-seeking} behavior, Reverse-KL allows the student to confidently capture the teacher's primary intent while retaining sufficient stochasticity to sample diverse, valid actions. This elegantly prevents both the entropy explosion of Forward-KL and the collapse of Hard-CE, ensuring highly stable updates that preserve pre-trained generalist capabilities and gracefully mitigate catastrophic forgetting.

Furthermore, this distillation paradigm fundamentally decouples the computationally prohibitive process of RL exploration from the student's policy optimization. While relying on an expert teacher assumes its prior availability, high-performing experts are increasingly accessible through open-source checkpoints, proprietary APIs, or easily trained single-task policies. VLA-OPD capitalizes on these existing resources by providing a highly sample-efficient distillation pipeline. This enables the seamless transfer of robust behaviors from diverse teachers into new, upgraded, or unified generalist student backbones. Consequently, our framework establishes a scalable pathway for continuous foundation model development, effectively circumventing the severe costs and instabilities associated with training VLA policies from scratch via online RL.

Our main contributions are summarized as follows:
\begin{itemize}
    \item We propose VLA-OPD, a unified post-training framework that bridges SFT and RL. By leveraging dense, token-level supervision on self-generated trajectories, it effectively resolves the exposure bias of SFT and the sample inefficiency of sparse-reward RL.
    
    \item We formulate a Reverse-KL distillation objective for VLA models. We demonstrate that its bounded mode-seeking property effectively filters out the teacher's epistemic uncertainty while maintaining action diversity, elegantly preventing both the entropy explosion of Forward-KL and the premature entropy collapse of Hard-CE.
    
    \item We provide a principled approach to mitigate catastrophic forgetting. By ensuring gradient updates remain grounded in the student's active policy manifold, VLA-OPD achieves a ``gentle'' alignment that preserves pre-trained generalist capabilities.
    
    \item Extensive evaluations across LIBERO and RoboTwin2.0 benchmarks demonstrate that VLA-OPD achieves superior robustness and success rates compared to SFT, while requiring substantially fewer training steps than on-policy RL baselines.
\end{itemize}

\vspace{-1mm}

%% file: sections/preliminaries.tex
\section{Preliminaries}
\label{sec:preliminaries}

In this section, we formalize the VLA training problem and briefly review the two dominant paradigms: Supervised Fine-Tuning (SFT) and Online Reinforcement Learning (RL).

\subsection{Problem Formulation}
We formulate the robotic manipulation task as a Markov Decision Process (MDP) defined by the tuple $(\mathcal{S}, \mathcal{A}, \mathcal{T}, r, \gamma)$. At each timestep $t$, the VLA agent observes a state $s_t \in \mathcal{S}$ (comprising visual observations and language instructions) and predicts an action $a_t \in \mathcal{A}$. The goal is to learn a policy $\pi_\theta(a_t | s_t)$ that maximizes the success rate of the task.

\subsection{Supervised Fine-Tuning (SFT)}
Standard VLA training typically starts with SFT on a static dataset of expert demonstrations $\mathcal{D}_{demo} = \{(\tau_i)\}$. The objective is to maximize the log-likelihood of the expert actions:
\begin{equation}
    \mathcal{L}_{SFT}(\theta) = - \mathbb{E}_{(s, a) \sim \mathcal{D}_{demo}} \left[ \log \pi_\theta(a | s) \right].
\end{equation}
While SFT provides dense supervision, it is \textit{off-policy}: the policy is trained on expert states but evaluated on student-induced states. This discrepancy leads to the \textit{distribution shift} problem discussed in Sec.~\ref{sec:intro}.

\subsection{Online RL with Sparse Outcome Rewards}
\label{sec:rl_prelim}

To address the limitations of SFT, researchers have increasingly turned to Online Reinforcement Learning. Drawing inspiration from its immense success in enhancing the reasoning capabilities of Large Language Models (LLMs)~\cite{guo2025deepseek}, Group Relative Policy Optimization (GRPO) has recently emerged as the promising approach for VLA post-training~\cite{li2025simplevla,zang2025rlinf}.

The widespread adoption of GRPO in the VLA domain stems from its architectural efficiency. By computing advantages via group-based relative normalization, GRPO eliminates the need for a separate value network (Critic). This significantly reduces memory overhead, making it uniquely suitable for fine-tuning large-scale vision-language backbones where maintaining a Critic is prohibitively expensive.

Formally, for an observation $s$, the policy samples a group of $G$ trajectories $\{\tau_1, \dots, \tau_G\}$. The optimization objective is to maximize the expected outcome reward:
\begin{equation}
    \mathcal{J}_{RL}(\theta) = \mathbb{E}_{s \sim \mathcal{D}, \tau \sim \pi_{\theta_{old}}} \left[ \frac{1}{G} \sum_{i=1}^G \min \left( \frac{\pi_\theta(\tau_i)}{\pi_{\theta_{old}}(\tau_i)} \hat{A}_i, \text{clip}\left(\frac{\pi_\theta(\tau_i)}{\pi_{\theta_{old}}(\tau_i)}, 1-\epsilon, 1+\epsilon\right) \hat{A}_i \right) \right],
    \label{eq:rl_objective}
\end{equation}
where $\epsilon$ is the clipping parameter, and $\hat{A}_i$ represents the advantage derived from a sparse outcome reward $R(\tau) \in \{0, 1\}$ (indicating task success or failure) relative to the group average.

However, despite its provenance from LLMs and architectural efficiency, applying GRPO to robotics introduces a unique challenge: feedback sparsity. Unlike reasoning tasks where intermediate steps might have clearer structure, robotics tasks often provide a binary signal only upon completion. This lack of granular supervision leads to high variance in optimization and severe sample inefficiency, necessitating prohibitively large amounts of interaction data to learn effective manipulation policies.

%% file: sections/method.tex
\section{Methodology}
\label{sec:method}

\begin{figure*}[t]
    \centering
    \includegraphics[width=\textwidth]{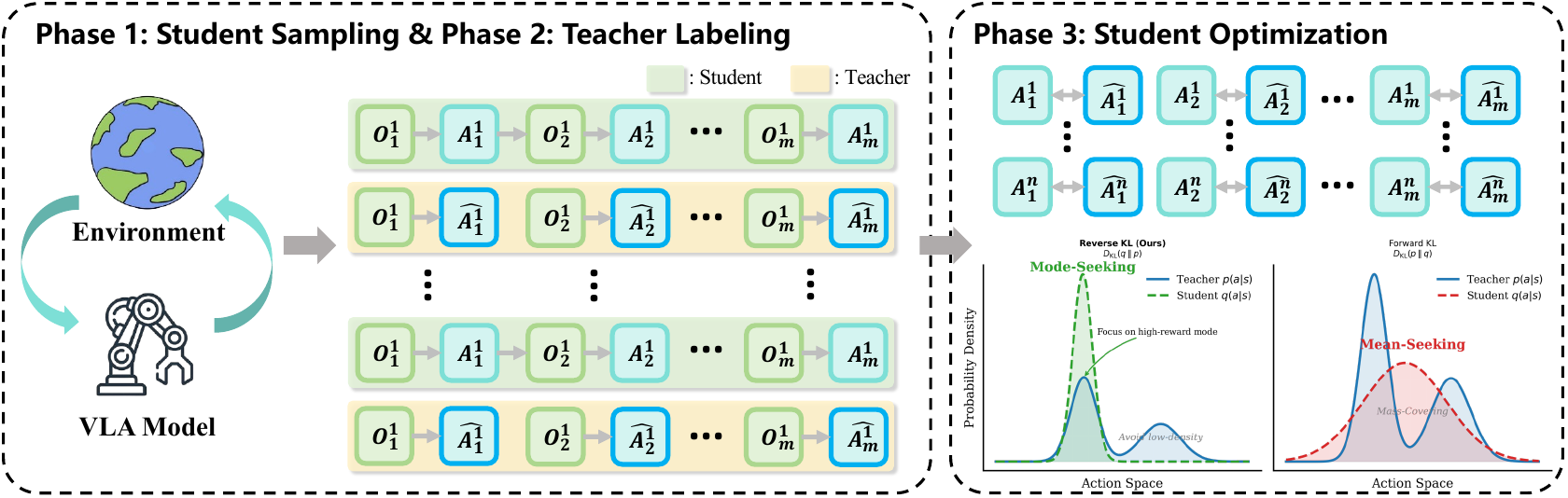}
    \vspace{-4mm}
    \caption{\textbf{Overview of VLA-OPD.} Our framework unifies offline SFT and online RL through three phases. 
    \textbf{Phase 1 (Student Sampling):} The student VLA policy interacts with the environment to collect on-policy trajectory rollouts ($O \rightarrow A \rightarrow O$). 
    \textbf{Phase 2 (Teacher Labeling):} For each state visited by the student, a frozen expert teacher provides dense, token-level action labels ($\widehat{A}$) without executing them in the environment. 
    \textbf{Phase 3 (Student Optimization):} The student is optimized against the teacher's distribution via a Reverse-KL objective. Unlike standard Forward-KL (bottom right) which induces mass-covering and entropy explosion, our Reverse-KL formulation (bottom left) promotes a \textbf{mode-seeking} behavior, effectively filtering out the teacher's out-of-distribution uncertainty and focusing on high-reward actions.}
    \label{fig:main_method}
    \vspace{-2mm}
\end{figure*}

In this section, we present \textbf{VLA-OPD}, a unified post-training framework designed to align Vision-Language-Action models efficiently and robustly. Our approach is motivated by the observation that standard SFT lacks the ability to recover from self-induced compounding errors, while on-policy RL suffers from feedback sparsity. To address these limitations simultaneously, VLA-OPD reformulates the alignment process as \textit{dense supervision on self-generated trajectories}.

We begin with a high-level overview of the framework (Sec.~\ref{sec:overview}). We then detail the on-policy sampling mechanism that addresses distribution shift (Sec.~\ref{sec:sampling}) and the teacher-guided dense supervision strategy that ensures sample efficiency (Sec.~\ref{sec:supervision}). Finally, we analyze our Reverse-KL optimization objective and its inherent mode-seeking properties (Sec.~\ref{sec:objective}).

\subsection{Framework Overview}
\label{sec:overview}

\begin{algorithm}[t]
\caption{VLA-OPD Training Procedure}
\label{alg:vla_opd}
\begin{algorithmic}[1]
\Require Student Policy $\pi_\theta$ (initialized from 1-traj SFT), Teacher Policy $\pi_{tea}$
\Require Group Size $G$, Learning Rate $\alpha$, Dataset $\mathcal{D}_{prompt}$

\State Initialize iteration counter $k \leftarrow 0$
\While{not converged}
    \State Sample a batch of prompts $\{o_j\}$ from $\mathcal{D}_{prompt}$
    
    \For{each prompt $o$ in batch}
        \State \textcolor{gray}{// Phase 1: Group Sampling (On-Policy)}
        \State Generate $G$ trajectories $\{\tau_1, \dots, \tau_G\}$ using student $\pi_{\theta}(\cdot|o)$
        
        \For{each trajectory $\tau_i$ in group}
            \State \textcolor{gray}{// Phase 2: Dense Teacher Labeling}
            \For{each timestep $t$ in $\tau_i$}
                \State Query student logits $\pi_\theta(\cdot | s_{t,i})$ and teacher logits $\pi_{tea}(\cdot | s_{t,i})$
                \State \textcolor{gray}{// Compute Intrinsic Reward (Negative Reverse-KL)}
                \State $r_t = - \left( \log \pi_\theta(a_{t,i}|s_{t,i}) - \log \pi_{tea}(a_{t,i}|s_{t,i}) \right)$
            \EndFor
        \EndFor
    \EndFor
    
    \State \textcolor{gray}{// Phase 3: Optimization (Group-Based Policy Gradient)}
    \State Estimate gradients averaged over group size $G$:
    \State $\nabla J \approx \frac{1}{B \times G} \sum_{j} \sum_{i=1}^G \sum_{t} \nabla_\theta \log \pi_\theta(a_{t,i}|s_{t,i}) \cdot r_t$
    \State Update parameters: $\theta \leftarrow \theta + \alpha \nabla J$
\EndWhile
\end{algorithmic}
\end{algorithm}

As illustrated in Figure~\ref{fig:main_method} and detailed in Algorithm~\ref{alg:vla_opd}, VLA-OPD operates as an iterative, closed-loop process involving two distinct policy roles.
The \textbf{Teacher Policy} ($\pi_{tea}$) is a robust expert model (e.g., trained via RL) that remains frozen during distillation. It acts as a reference oracle, providing dense supervision signals that enable the student to learn optimal recovery behaviors even in states not covered by the original expert demonstrations. 
The \textbf{Student Policy} ($\pi_{\theta}$) is the target VLA model being trained, typically initialized from a base checkpoint (e.g., via offline SFT). By interacting with the environment on-policy, the student collects trajectories and is continuously updated to align with the teacher's robust distribution.

The goal of VLA-OPD is to efficiently transfer the robustness of $\pi_{tea}$ to the brittle $\pi_{\theta}$. As depicted in Figure~\ref{fig:main_method}, the training cycle consists of three phases:

\begin{itemize}
    \item \textbf{Phase 1: On-Policy Sampling (Exploration).} 
    The student $\pi_\theta$ generates trajectories $\mathcal{T}_{student}$ in the environment. Since the student is trained on limited data, it frequently encounters out-of-distribution states. This explicitly triggers the \textit{distribution shift}, exposing the model to the boundaries of its capabilities.
    
    \item \textbf{Phase 2: Dense Teacher Labeling (Correction).} 
    Instead of waiting for sparse outcome rewards, we query the frozen teacher $\pi_{tea}$. For every state $s_t$ visited by the student, the teacher provides its action logits. This acts as a dense guiding signal, effectively injecting an optimal recovery prior to correct the student's deviations in unfamiliar states.
    
    \item \textbf{Phase 3: Mode-Seeking Optimization (Update).}
    The student is updated via on-policy policy-gradient using the token-level Reverse-KL reward, which is equivalent to minimizing the divergence from the teacher on student-visited states.
\end{itemize}

\subsection{On-Policy Trajectory Sampling}
\label{sec:sampling}

A fundamental limitation of the SFT initialization is the \textit{distribution shift} problem. Since the student model is trained on a highly restricted set of expert states $\mathcal{D}_{expert}$, it lacks knowledge of how to behave in states outside this narrow manifold. During evaluation, minor execution errors accumulate, driving the agent into unfamiliar states where its policy is undefined, leading to catastrophic failure.

To address this, VLA-OPD discards the static offline dataset after initialization and switches to dynamic \textbf{on-policy sampling}. At each training iteration $k$, we collect a batch of trajectories $\mathcal{D}_k$ by executing the current student policy $\pi_{\theta_k}$ in the environment:
\begin{equation}
    \mathcal{D}_k = \{ \tau \mid \tau = (s_0, a_0, s_1, a_1, \dots, s_T) \}, \quad \text{where } a_t \sim \pi_{\theta_k}(\cdot | s_t), \ s_{t+1} \sim \mathcal{P}(\cdot | s_t, a_t).
    \label{eq:rollout}
\end{equation}

Crucially, the states $s_t$ in $\mathcal{D}_k$ are drawn from the student's \textit{induced distribution} $d^{\pi_{\theta_k}}$, rather than the expert distribution. 
For a brittle 1-traj student, this sampling process is primarily driven by the need for active correction. Because the student frequently deviates from the expert's path, VLA-OPD explicitly captures these ``failure states'' ($s_{err}$) in $\mathcal{D}_k$. By subsequently training on these samples (as detailed in Sec.~\ref{sec:supervision}), we effectively convert the ``unknown'' out-of-distribution regions into ``known'' training data, transforming the alignment problem from passive imitation to active correction.

Beyond robust correction, this on-policy formulation provides a principled mechanism for mitigating catastrophic forgetting. Standard SFT is inherently off-policy, forcing the model to fit a fixed, disjoint target distribution, which necessitates aggressive parameter shifts that overwrite pre-trained generalist knowledge. In contrast, VLA-OPD ensures that gradient updates remain anchored to the student's current behavioral manifold. By distilling knowledge strictly on trajectories the student naturally visits, our approach achieves a ``gentle'' alignment that effectively preserves the backbone's pre-trained capabilities.

\subsection{Dense Teacher Supervision}
\label{sec:supervision}

Instead of relying on sparse outcome rewards that suffer from severe credit assignment issues, VLA-OPD leverages the robust teacher $\pi_{tea}$ to provide dense, token-level supervision. For every timestep $t$ in a student-generated trajectory $\tau \in \mathcal{D}_k$, we query the teacher to obtain the target action distribution:
\begin{equation}
    q_t(a) = \pi_{tea}(a | s_t).
\end{equation}
This dense signal offers two critical advantages. First, it converts the delayed RL problem into an immediate supervised signal, drastically accelerating convergence. Second, by labeling the student's on-policy states, including out-of-distribution deviations, the teacher imparts structural knowledge~\cite{hinton2014dark} on optimal recovery behaviors. However, standard distillation across the entire distribution can be detrimental when the teacher is uncertain. To selectively leverage this knowledge while filtering out high-entropy noise, we introduce a mode-seeking objective, which we discuss in detail in Section~\ref{sec:objective}.

\subsection{Optimization Objective and Analysis}
\label{sec:objective}

To align the student policy $\pi_\theta$ with the robust teacher $\pi_{tea}$, VLA-OPD employs an optimization strategy grounded in minimizing the \textbf{Reverse Kullback-Leibler (KL) divergence}.

\noindent\textbf{Reverse-KL as Dense Reward.}
Our goal is to minimize the divergence between the student and teacher distributions on student-generated trajectories. To formulate this as a reinforcement learning problem, we define the objective as maximizing the \textbf{negative} Reverse-KL divergence:
\begin{equation}
    \max_\theta \mathcal{J}(\theta) = \mathbb{E}_{s \sim \pi_\theta} \left[ - D_{KL}(\pi_\theta(\cdot|s) || \pi_{tea}(\cdot|s)) \right].
\end{equation}
At the token level, this translates to an intrinsic reward $r_t^{OPD}$ defined as the negative log-ratio of the probabilities:
\begin{equation}
    r_t^{OPD}(s_t, a_t) = - \left( \log \pi_\theta(a_t | s_t) - \log \pi_{tea}(a_t | s_t) \right) = - \log \frac{\pi_\theta(a_t | s_t)}{\pi_{tea}(a_t | s_t)}.
    \label{eq:kl_reward}
\end{equation}
Intuitively, this reward acts as a penalty: the student receives a higher reward (closer to 0) when its action distribution matches the teacher's, and a large negative penalty when it deviates significantly. Crucially, when computing the policy gradient update, we apply a \texttt{stop\_gradient} operation to the student's log-probability term $\log \pi_\theta(a_t | s_t)$ within the reward calculation. 

\noindent\textbf{Theoretical Analysis: Mode-Seeking vs. Mode-Covering.}
The choice of divergence direction fundamentally alters the optimization dynamics, particularly in OOD states where the teacher may exhibit high epistemic uncertainty (i.e., flat, high-entropy distributions).

\begin{itemize}
    \item \textbf{Forward KL (Teacher-Forced):} Standard SFT minimizes $D_{\mathrm{KL}}(\pi_{\mathrm{tea}} \parallel \allowbreak \pi_\theta)$. Its gradient is estimated over teacher samples ($\mathbb{E}_{a \sim \pi_{\mathrm{tea}}}$), effectively forcing the student to cover the teacher's entire support (\textbf{mode-covering}). In OOD states, this compels the student to mimic the teacher's hesitation and high entropy, leading to the \textbf{entropy explosion} phenomenon.
    
    \item \textbf{Hard-CE (Argmax Matching):} A common alternative in on-policy settings (e.g., standard DAgger) is minimizing Cross-Entropy against the teacher's top-1 action. Mathematically, this discards the teacher's soft probabilities ("dark knowledge"). When the teacher's argmax oscillates at multi-modal decision boundaries, Hard-CE forces the student to violently track these rigid targets, causing \textbf{premature entropy collapse} and depriving the student of the action diversity necessary for robust exploration.

    \item \textbf{Reverse-KL (Bounded Mode-Seeking):} In contrast, VLA-OPD minimizes $D_{\mathrm{KL}}(\pi_\theta \parallel \pi_{\mathrm{tea}})$. Due to the \textit{zero-forcing} property of Reverse-KL, as long as the student's chosen action falls within the teacher's acceptable probability mass, it is not penalized for ignoring other potential actions. This induces a \textbf{mode-seeking} behavior that elegantly avoids both extremes: it filters out the teacher's tail uncertainty (preventing entropy explosion) while retaining sufficient stochasticity within the valid modes (preventing premature entropy collapse).

\end{itemize}

We validate these distinct optimization dynamics in our ablation studies (Section~\ref{sec:ablation}). As vividly demonstrated in Figure~\ref{fig:kl_ablation_comparison}, Forward-KL indeed suffers from severe entropy explosion, Hard-CE experiences premature entropy collapse leading to sub-optimal plateaus, whereas our Reverse-KL maintains a healthy, bounded entropy that translates to the highest and most stable task success rate.

\noindent\textbf{Group-Based Gradient Estimation.}
To reduce the high variance typically associated with on-policy gradients, we adopt a group sampling strategy. Specifically, for each instruction $s$, we sample a group of $G$ trajectories $\{\tau_1, \dots, \tau_G\}$ from the current student policy $\pi_\theta$. We estimate the policy gradient by averaging over this group:
\begin{equation}
    \nabla_\theta \mathcal{J}(\theta) \approx \frac{1}{G} \sum_{i=1}^G \sum_{t=0}^T \nabla_\theta \log \pi_\theta(a_{t,i} | s_{t,i}) \cdot r_t^{OPD}(s_{t,i}, a_{t,i}).
    \label{eq:gradient}
\end{equation}
Unlike standard GRPO which computes advantages via outcome reward normalization, our method uses the raw Reverse-KL reward directly as the advantage signal, ensuring consistent convergence toward the teacher's optimal mode.

%% file: sections/experiments.tex
\section{Experiments}
\label{sec:experiments}

Our experimental design addresses several fundamental research questions regarding the efficiency, efficacy, and stability of On-Policy Distillation (OPD) for robotic manipulation:

\begin{enumerate}
    \item \textbf{Training Efficiency:} Does dense teacher supervision enable significantly faster convergence compared to standard online reinforcement learning (e.g., GRPO), which relies on sparse reward exploration?
    \item \textbf{Policy Efficacy:} To what extent can VLA-OPD recover robust performance from a sub-optimal base policy, whether constrained by extreme data scarcity (e.g., 1-traj SFT) or morphological complexity (e.g., dual-arm coordination), and bridge the gap to an expert teacher?
    \item \textbf{Catastrophic Forgetting:} Does on-policy distillation better preserve pre-trained generalist capabilities than offline SFT, while improving performance on the target tasks?
    \item \textbf{Ablation and Design Choices:} How do the core algorithmic components of VLA-OPD—such as the alignment objective—contribute to the overall stability and success rate of the policy?
\end{enumerate}

\subsection{Experimental Setup}
\label{sec:setup}

\noindent\textbf{Benchmarks and Protocols.} We evaluate across two distinct domains to test data-scarce generalization and complex coordination. \textbf{(1) LIBERO}~\cite{liu2023libero}: We utilize four suites (Spatial, Object, Goal, Long) for single-arm manipulation. To evaluate under extreme data scarcity, student models are initialized via \textbf{1-traj SFT} (one demo per task). \textbf{(2) RoboTwin2.0}~\cite{chen2025robotwin}: We select four representative tasks requiring complex dual-arm coordination. Given its inherent difficulty, students are initialized via \textbf{1,000-traj SFT} per task; the sub-optimal base policy provides an ideal testbed for our distillation framework.

\noindent\textbf{Teacher and Baselines.} We employ SimpleVLA-RL~\cite{li2025simplevla} as our teacher $\pi_{tea}$ (Performance Oracle). We compare against: \textbf{(1) Student Init. (Lower Bound):} The base OpenVLA-OFT~\cite{kim2025finetuningvisionlanguageactionmodelsoptimizing} models before distillation (1-traj and 1,000-traj SFT for LIBERO and RoboTwin2.0, respectively). \textbf{(2) Online RL:} GRPO with sparse rewards. \textbf{(3) Offline SFT:} Models fine-tuned on full expert datasets.

\subsection{Main Results: Efficiency and Efficacy}
\label{sec:efficiency}

We compare our method against the baseline GRPO. To ensure a fair comparison, following SimpleVLA-RL~\cite{li2025simplevla}, we set the batch size to 64 and the group size to $G=8$ across main experiments. Our approach is evaluated in two settings: (1) \textbf{Ours (Distill)}, which uses only the distillation process, and (2) \textbf{Ours (Distill + GRPO)}, which further fine-tunes the distilled model using GRPO.

\begin{figure*}[t]
    \centering
    \begin{subfigure}[b]{0.48\textwidth}
        \centering
        \includegraphics[width=\textwidth]{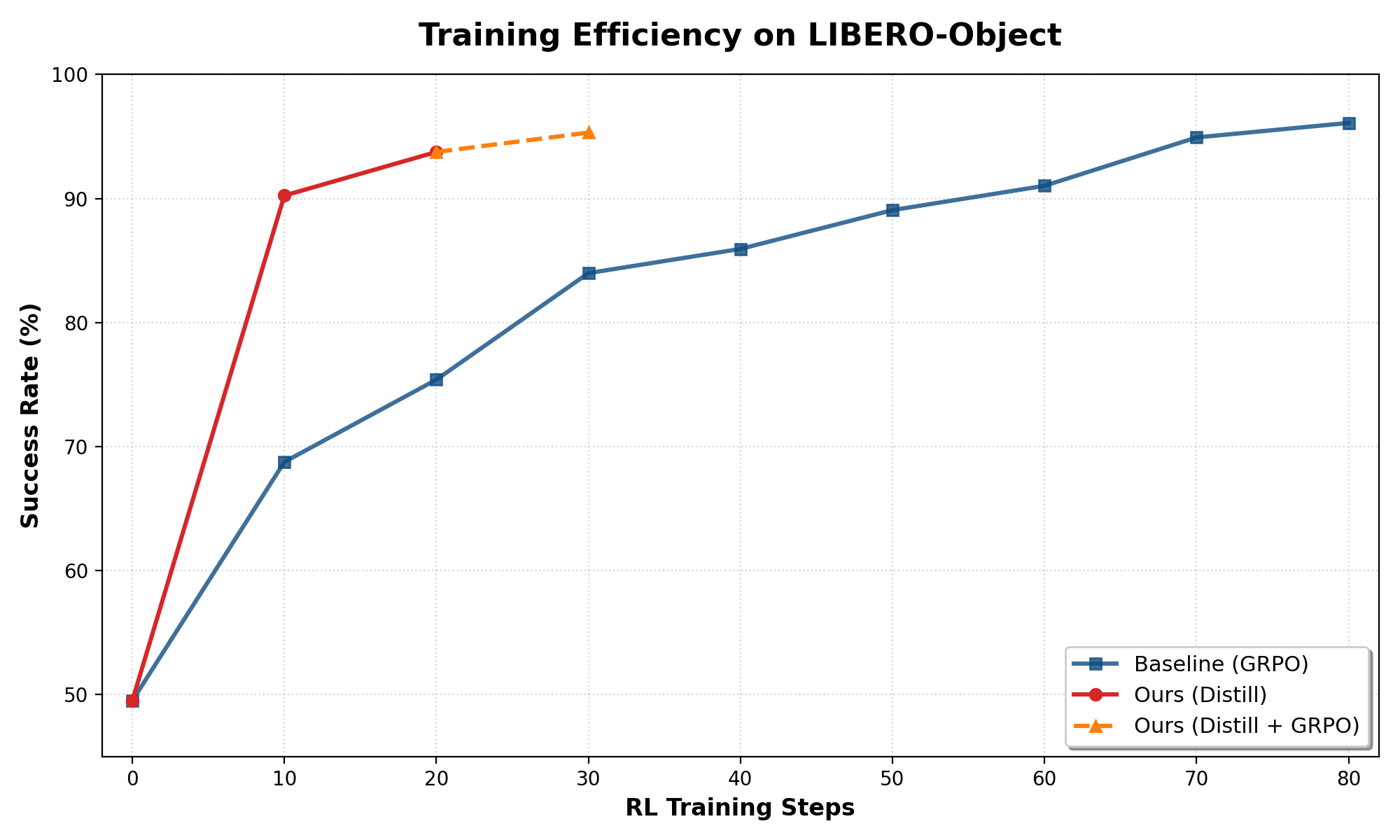}
        \caption{Training Efficiency on LIBERO-Object}
        \label{fig:eff_object}
    \end{subfigure}
    \hfill
    \begin{subfigure}[b]{0.48\textwidth}
        \centering
        \includegraphics[width=\textwidth]{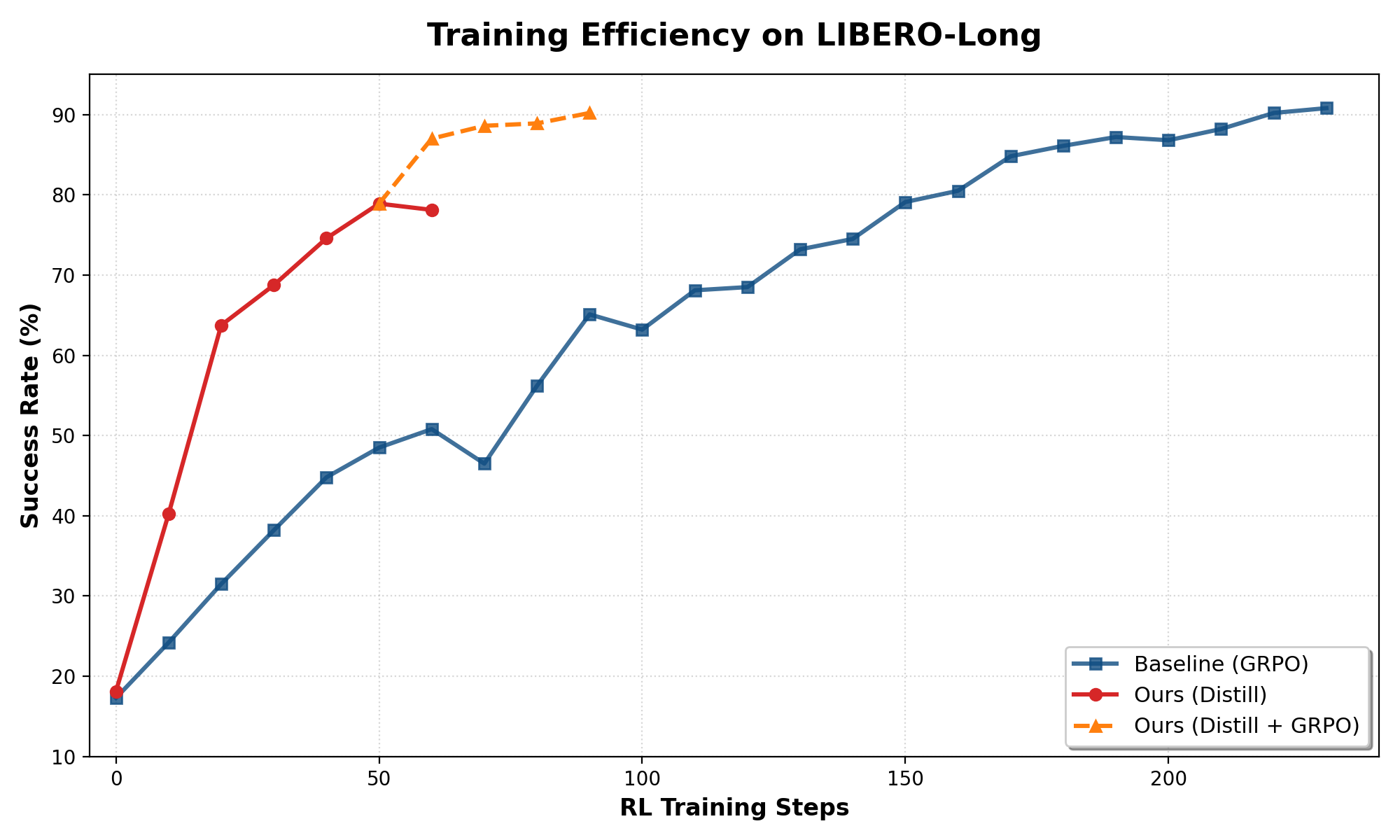}
        \caption{Training Efficiency on LIBERO-Long}
        \label{fig:eff_long}
    \end{subfigure}
    
    \caption{\textbf{Training Efficiency Comparison.} 
    We compare our method with the baseline GRPO across two benchmarks. 
    The red line (\textbf{Ours (Distill)}) demonstrates superior sample efficiency in the early stages, achieving high success rates with significantly fewer steps. 
    The dashed orange line (\textbf{Ours (Distill + GRPO)}) shows that further RL fine-tuning breaks the performance bottleneck, surpassing the baseline's final convergence. 
    Notably, on LIBERO-Long (b), our method achieves near 80\% success rate in just 50 steps, whereas the baseline requires over 150 steps, representing a \textbf{3$\times$ speedup}.}
    \label{fig:training_efficiency}
\end{figure*}

As shown in Figure~\ref{fig:training_efficiency}, our method demonstrates significant advantages in both data efficiency and stability:

\begin{itemize}
    \item \textbf{Rapid Convergence:} On LIBERO-Object (Figure~\ref{fig:eff_object}), our distillation method achieves over 90\% success rate within just 10 steps, exhibiting a ``vertical'' takeoff compared to the gradual climb of the baseline. Similarly, on LIBERO-Long (Figure~\ref{fig:eff_long}), we reach comparable performance to the baseline's 150-step result in only 50 steps.
    
    \item \textbf{Breaking Performance Ceilings:} While distillation alone provides a strong ``warm start'', the combination with GRPO (dashed orange line) further pushes the performance boundaries, achieving state-of-the-art results (over 95\% on Object and 90\% on Long).
    
    \item \textbf{Stability:} Unlike the baseline GRPO, which suffers from severe fluctuations (zig-zag patterns) particularly visible in the LIBERO-Long task, our training curves are remarkably smooth, indicating a more robust and stable optimization process.
\end{itemize}

\begin{table}[H]
    \centering
    \scriptsize
    \setlength{\tabcolsep}{4pt}
    \caption{\textbf{Main Results: Policy Efficacy on LIBERO.} We report the success rate (\%). 
    We compare VLA-OPD (trained on 1-traj) against two groups of baselines: 
    (1) \textbf{Full-Dataset Methods}: Models trained on the complete expert dataset (50 demos/task), representing the data-abundant upper bound.
    (2) \textbf{Data-Scarce Methods}: Models trained on the same 1-traj split.
    \textbf{VLA-OPD} achieves performance comparable to Full-Dataset methods, significantly outperforming other data-scarce approaches.}
    \label{tab:main_results}
    \begin{tabular}{lccccc}
        \toprule
        \textbf{Method} & \textbf{Spatial} & \textbf{Object} & \textbf{Goal} & \textbf{Long} & \textbf{Avg.} \\
        \midrule
        \textit{Teacher (Reference)} & & & & & \\
        SimpleVLA-RL~\cite{li2025simplevla} & 94.2 & 96.1 & 94.6 & 90.7 & 93.9 \\
        \midrule
        \textit{Full-Dataset Methods (50-traj)} & & & & & \\
        Octo~\cite{team2024octo} & 78.9 & 85.7 & 84.6 & 51.1 & 75.1 \\
        OpenVLA~\cite{kim2024openvla} & 84.7 & 88.4 & 79.2 & 53.7 & 76.5 \\
        Nora~\cite{hung2025nora} & 92.2 & 95.4 & 89.4 & 74.6 & 87.9 \\ 
        $\pi_0$ + FAST~\cite{pertsch2025fast} & 96.4 & 96.8 & 88.6 & 60.2 & 85.5 \\
        \midrule
        \textit{Data-Scarce Methods (1-traj)} & & & & & \\
        OpenVLA-OFT~\cite{kim2025finetuningvisionlanguageactionmodelsoptimizing} (Student Init.) & 63.6 & 54.9 & 59.6 & 17.3 & 48.9 \\
        \rowcolor{gray!15} 
        \textbf{VLA-OPD (Ours) (Distill)} & \textbf{84.3} & \textbf{93.8} & \textbf{92.5} & \textbf{78.9} & \textbf{87.4} \\
        \rowcolor{gray!15} 
        \textbf{VLA-OPD (Ours) (Distill + GRPO)} & \textbf{93.4} & \textbf{95.3} & \textbf{94.5} & \textbf{90.2} & \textbf{93.4} \\
        \bottomrule
    \end{tabular}
\end{table}

Beyond training efficiency, we evaluate the final policy efficacy across four LIBERO task suites. As shown in Table~\ref{tab:main_results}, VLA-OPD achieves striking performance improvements under the data-scarce (1-traj) setting. While the student initialization (OpenVLA-OFT) struggles with an average success rate of only 48.9\%, our pure distillation variant (\textbf{Ours (Distill)}) boosts the performance to 87.4\%, effectively matching or surpassing several full-dataset (50-traj) baselines like Octo~\cite{team2024octo} and OpenVLA~\cite{kim2024openvla}. Furthermore, by seamlessly integrating the distillation warm-start with subsequent RL fine-tuning (\textbf{Ours (Distill + GRPO)}), the student policy reaches an impressive 93.4\% average success rate. This nearly recovers the performance of the expert teacher (93.9\%) while bypassing the prohibitive exploration costs of training a VLA from scratch.

\noindent\textbf{Extension to Dual-Arm Manipulation (RoboTwin2.0).} 
To verify that VLA-OPD is not limited to single-arm tabletop tasks, we extend our evaluation to the RoboTwin2.0 benchmark, which features complex dual-arm coordination tasks. As shown in Table~\ref{tab:robotwin_selected}, despite being initialized with 1,000 demonstrations per task via single-task SFT (OpenVLA-OFT), the student still struggles significantly due to the increased morphological complexity, averaging only 45.2\% success. However, by applying our distillation framework (\textbf{Ours (Distill)}), the student's average success rate surges to 71.1\%, nearly matching the Teacher's performance (74.0\%) and substantially outperforming other baselines such as $\pi_0$~\cite{black2024pi0visionlanguageactionflowmodel} and RDT~\cite{liu2024rdt}. This confirms the efficacy and morphological generalization of VLA-OPD in highly complex environments.

\begin{table}[t] 
    \centering
    \footnotesize
    \caption{\textbf{Evaluation on Selected RoboTwin2.0 Tasks.} 
    Following the setting in~\cite{li2025simplevla}, models are initialized with full-dataset SFT (1000 demos/task).
    We report success rates (\%) on four representative tasks with varying horizon lengths.
    \textbf{VLA-OPD} consistently improves over the SFT initialization, especially in long-horizon tasks.}
    \label{tab:robotwin_selected}
    
    \resizebox{\linewidth}{!}{
    \begin{tabular}{lccccc}
        \toprule
        & \textbf{Short} & \textbf{Medium} & \textbf{Long} & \textbf{Long} & \\
        \textbf{Method} & \textit{Pick dual bottles} & \textit{Place Empty Cup} & \textit{Handover Block} & \textit{Stack Bowls Two} & \textbf{Avg.} \\
        \midrule
        \textit{Teacher (Reference)} & & & & & \\
        SimpleVLA-RL~\cite{li2025simplevla} & 68.3 & 94.2 & 57.8 & 75.8 & 74.0 \\
        \midrule
        \textit{Baselines} & & & & & \\
        $\pi_0$~\cite{black2024pi0visionlanguageactionflowmodel} & 50.0 & 60.0 & 39.0 & 53.0 & 50.5 \\
        RDT~\cite{liu2024rdt} & 18.0 & 42.0 & 26.0 & 42.0 & 32.0 \\
        \midrule
        OpenVLA-OFT~\cite{kim2025finetuningvisionlanguageactionmodelsoptimizing} (Student Init.) & 29.7 & 77.3 & 33.1 & 40.6 & 45.2 \\
        \rowcolor{gray!15} 
        \textbf{VLA-OPD (Ours) (Distill)} & \textbf{66.4} & \textbf{90.6} & \textbf{52.3} & \textbf{75.0} & \textbf{71.1} \\
        \bottomrule
    \end{tabular}
    }
\end{table}

\subsection{Mitigating Catastrophic Forgetting via On-Policy Alignment}
\label{sec:forgetting}

\begin{figure*}[t] 
    \centering
    \begin{subfigure}[b]{0.35\textwidth}
        \centering
        \includegraphics[width=\textwidth]{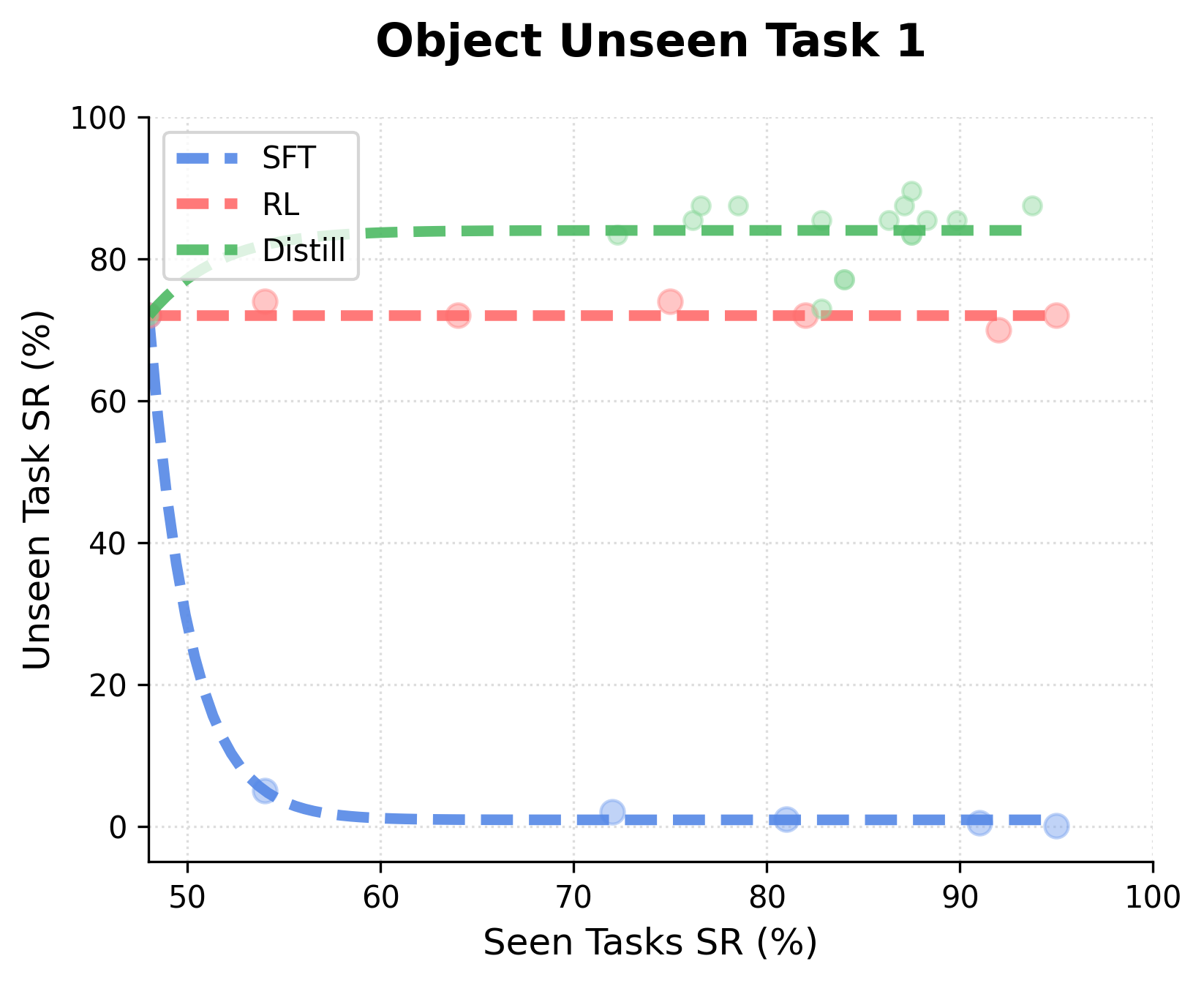}
        \label{fig:obj-unseen-1}
    \end{subfigure}
    \hspace{0.05\textwidth}
    \begin{subfigure}[b]{0.35\textwidth}
        \centering
        \includegraphics[width=\textwidth]{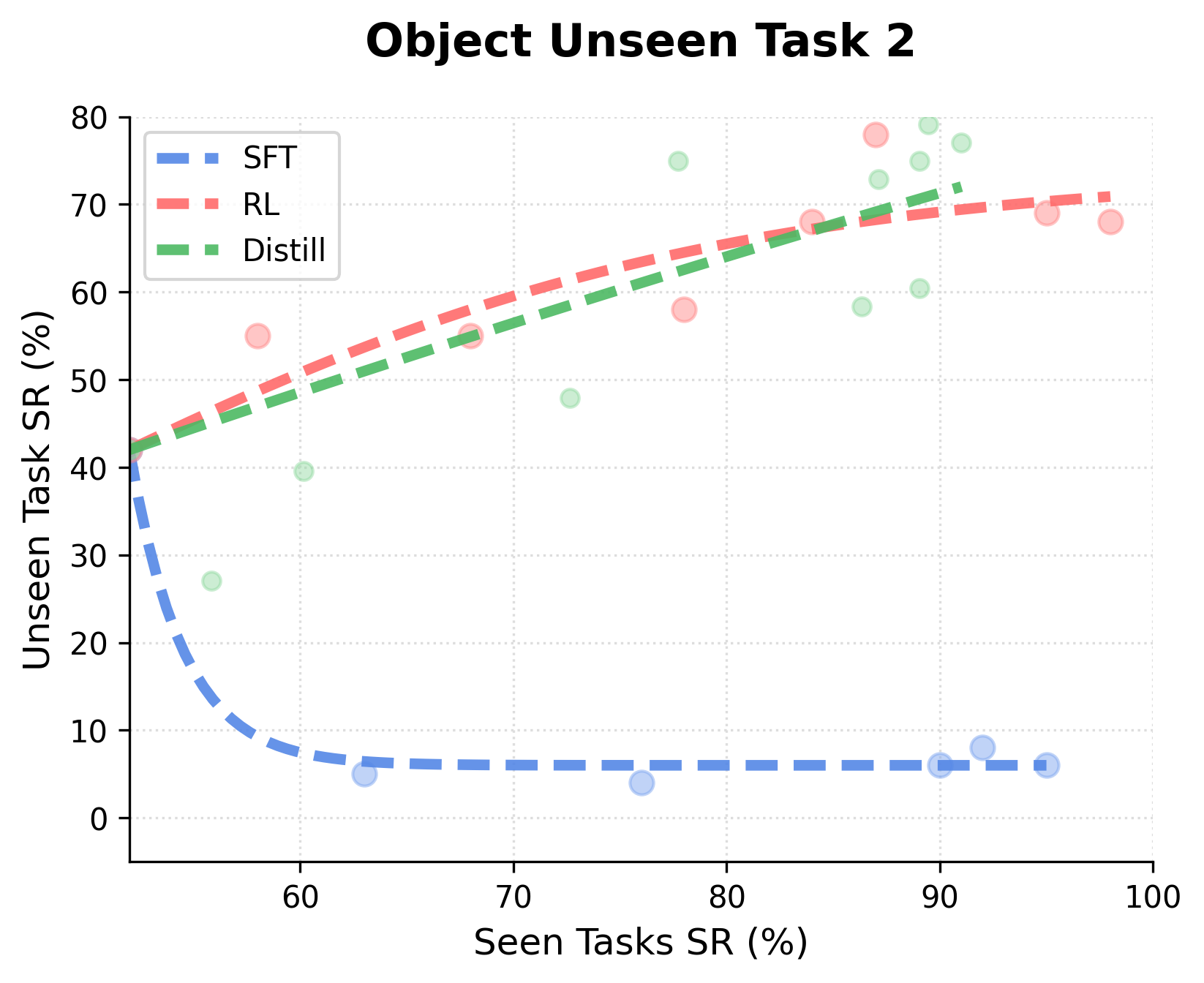}
        \label{fig:obj-unseen-2}
    \end{subfigure}
    
    \begin{subfigure}[b]{0.35\textwidth}
        \centering
        \includegraphics[width=\textwidth]{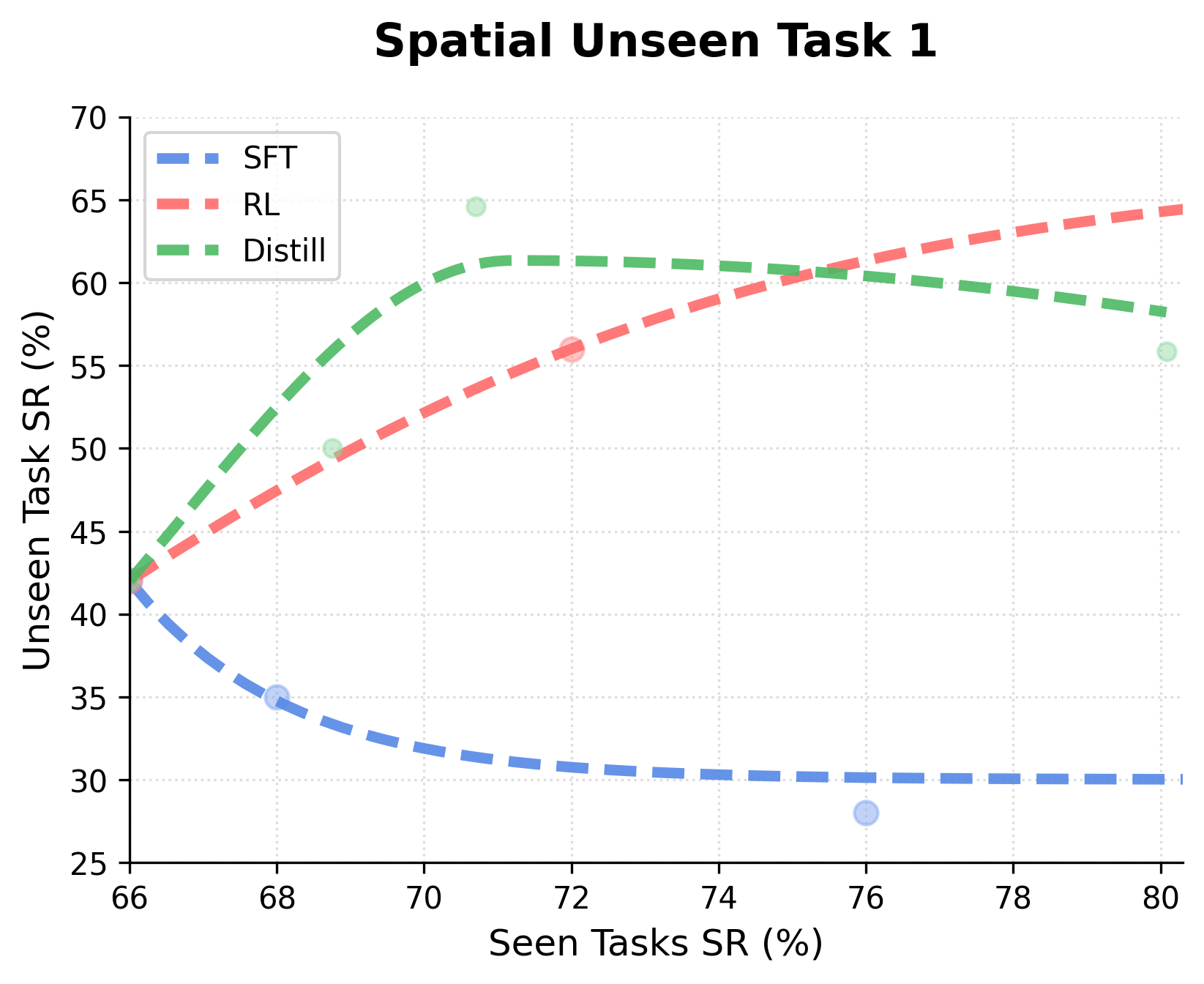}
        \label{fig:sp-unseen-1}
    \end{subfigure}
    \hspace{0.05\textwidth}
    \begin{subfigure}[b]{0.35\textwidth}
        \centering
        \includegraphics[width=\textwidth]{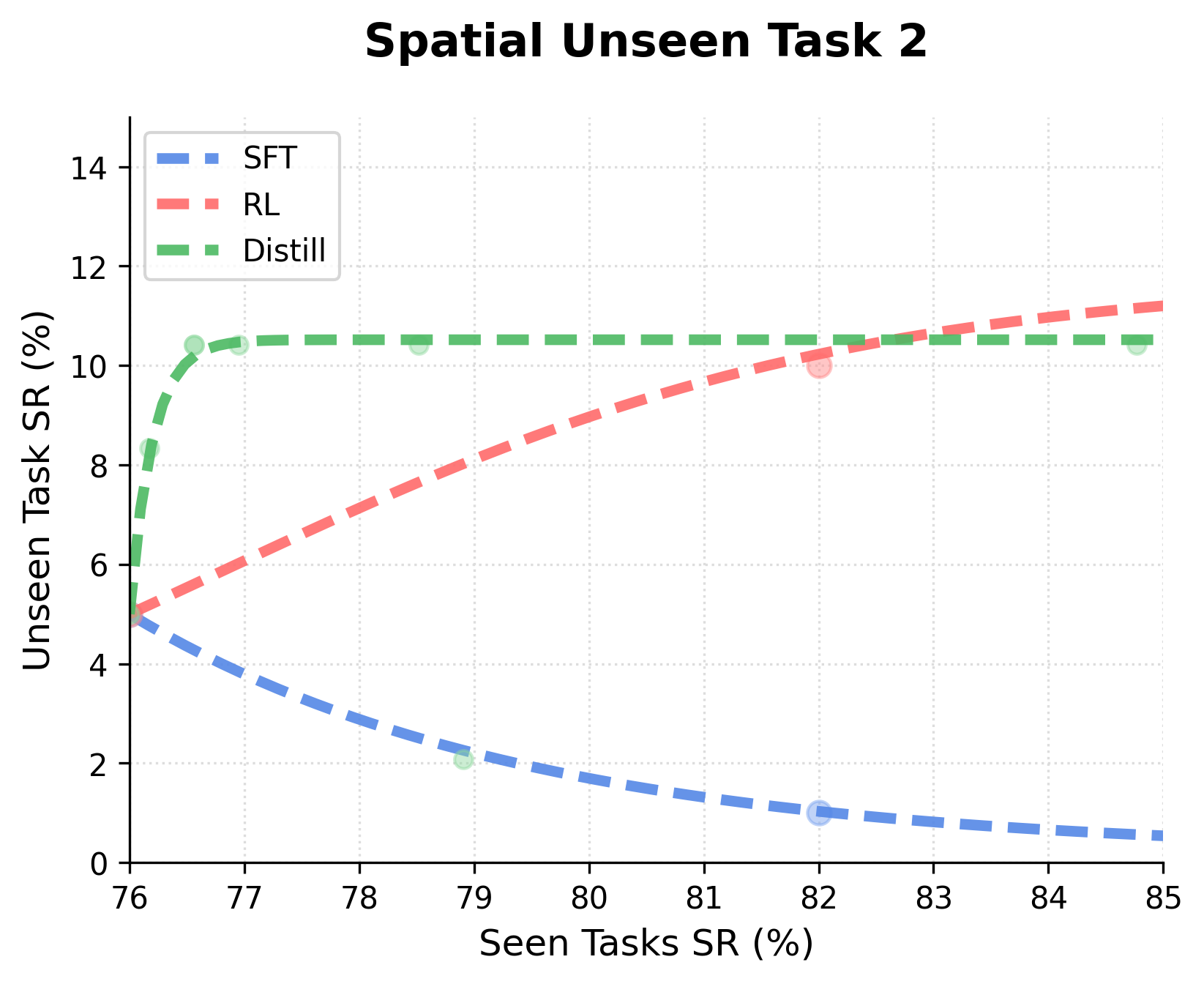}
        \label{fig:sp-unseen-2}
    \end{subfigure}

    \caption{\textbf{Seen--Unseen Trade-off for Forgetting Analysis.}
    Each point corresponds to a checkpoint during fine-tuning.
    The x-axis is the success rate on \textbf{seen (target)} tasks, and the y-axis is the success rate on a held-out \textbf{unseen} task.
    Offline SFT exhibits a strong collapse on unseen tasks as seen-task performance increases, while on-policy methods (RL and our distillation) better preserve unseen-task capability.}
    \label{fig:forgetting_tradeoff}
\end{figure*}

We analyze catastrophic forgetting via a \textbf{seen--unseen trade-off} (Figure~\ref{fig:forgetting_tradeoff}), fine-tuning on target (seen) tasks and evaluating on four held-out \textbf{unseen tasks} (two Object, two Spatial). The ideal upper-right region indicates strong target mastery alongside preserved general capabilities. 

Consistently, \textbf{offline SFT exhibits severe forgetting}: as seen-task success improves, unseen performance collapses—approaching zero for Object tasks and dropping substantially for Spatial tasks—confirming that optimizing solely on offline trajectories quickly erodes pre-trained skills due to distribution shifts. Conversely, methods leveraging \textbf{on-policy data} (RL and VLA-OPD) largely avoid this collapse. VLA-OPD matches or exceeds RL across multiple axes, achieving strong preservation on Object Task 1, remaining competitive on Object Task 2, and providing comparable retention on Spatial tasks.

This simultaneous mastery without degradation demonstrates VLA-OPD's suitability for continual learning. By safeguarding pre-trained knowledge via on-policy supervision, it offers a sustainable paradigm for lifelong robot learning without requiring the heavy replay buffers typically needed in offline adaptation. Ultimately, on-policy data keeps the model within a safer distributional neighborhood, effectively mitigating the catastrophic forgetting inherent to offline SFT.

\subsection{Ablation Studies}
\label{sec:ablation}
To validate the architectural design of VLA-OPD, we conduct ablation studies focusing on two critical components: the choice of alignment objective (Reverse-KL vs. Forward-KL vs. Hard-CE) and the impact of group sampling size. All ablation experiments are conducted with a fixed batch size of 32.

\begin{figure}[htbp]
    \centering
    \begin{subfigure}[b]{0.49\textwidth}
        \centering
        \includegraphics[width=\textwidth]{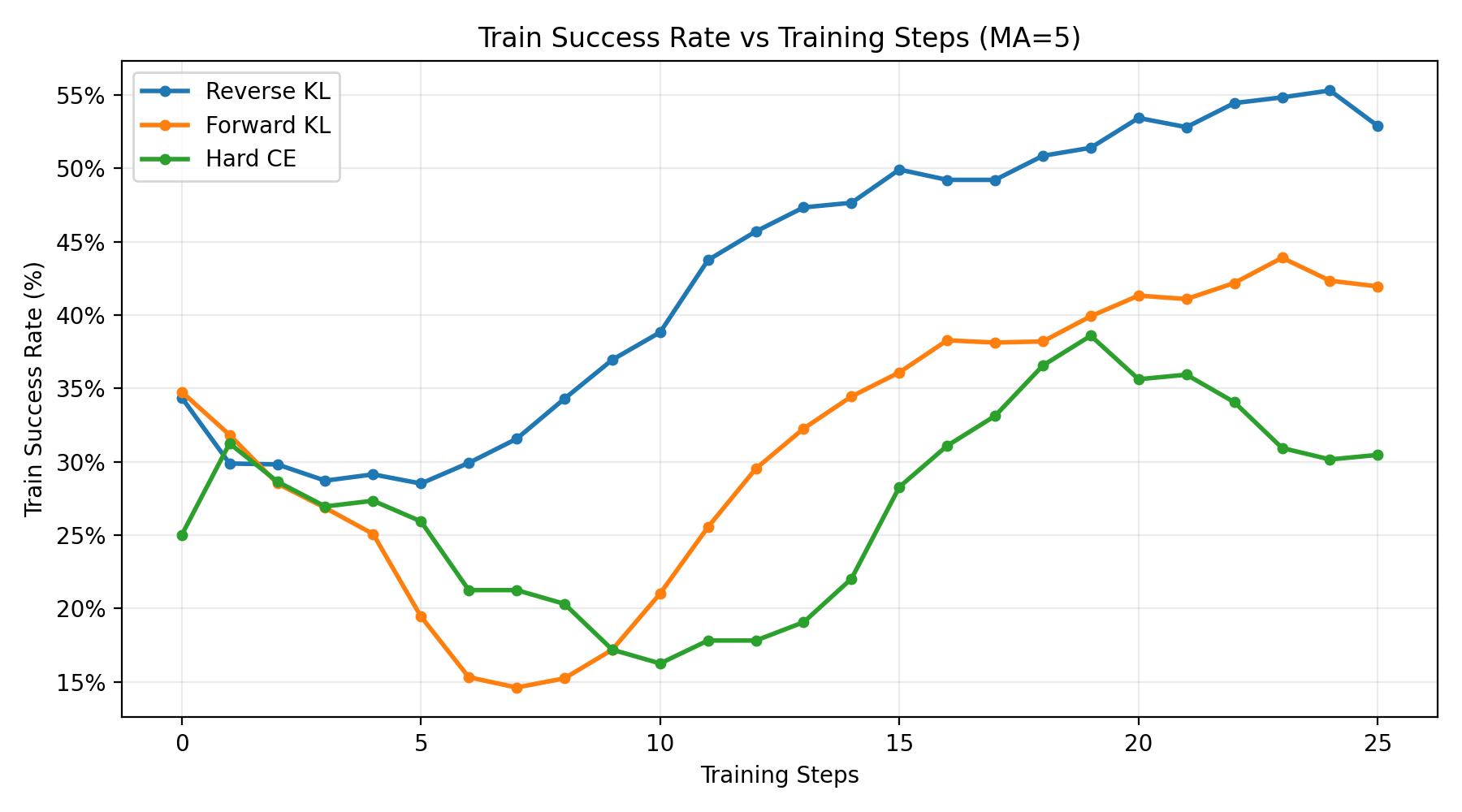} 
        \caption{Training Success Rate}
        \label{fig:kl_success_rate}
    \end{subfigure}
    \hfill
    \begin{subfigure}[b]{0.49\textwidth}
        \centering
        \includegraphics[width=\textwidth]{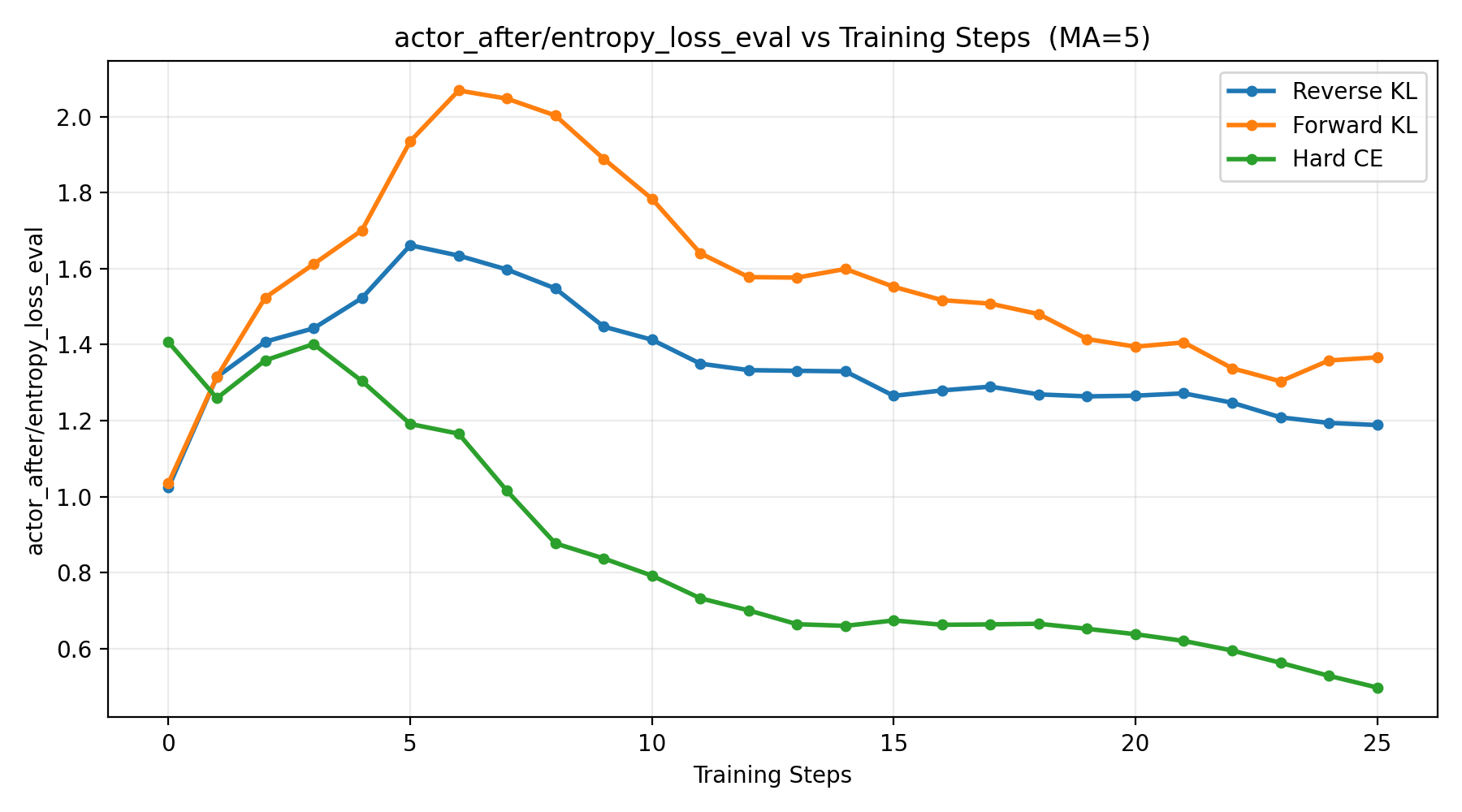} 
        \caption{Actor Entropy}
        \label{fig:kl_entropy}
    \end{subfigure}
    \caption{Ablation study comparing Reverse KL, Forward KL, and Hard CE in an on-policy distillation setting, evaluated on the RoboTwin2.0 \textit{Beat Block Hammer} task.
    (a) Forward KL suffers a severe early performance drop, and Hard CE plateaus at a suboptimal level, whereas Reverse KL shows steady and superior improvement. 
    (b) These performance differences correlate directly with entropy extremes: Forward KL induces \textbf{entropy explosion} (mode-covering), while Hard CE causes \textbf{premature entropy collapse} (loss of action diversity). In contrast, Reverse KL maintains a healthy, bounded entropy via \textbf{mode-seeking}, ensuring stable training.}
    \label{fig:kl_ablation_comparison}
\end{figure}

\noindent\textbf{1. Alignment Objective: Reverse KL vs. Forward KL vs. Hard CE.}
To verify the optimal alignment objective for our framework, we conduct a comparative experiment against standard Forward KL and Hard CE (e.g., standard DAgger). As shown in Figure~\ref{fig:kl_success_rate}, while Reverse KL leads to a steady and robust improvement in task success rate, the alternatives struggle significantly. Forward KL suffers from a severe ``performance valley'' where the success rate drops by more than 50\% during early stages. Meanwhile, Hard CE fails to recover effectively, ultimately plateauing at the lowest success rate.

These performance disparities are intrinsically linked to the optimization dynamics in Out-Of-Distribution (OOD) states, which are vividly reflected in the actor's entropy (Figure~\ref{fig:kl_entropy}). During early training, the on-policy student frequently visits OOD states where the teacher exhibits high epistemic uncertainty. 
The \textbf{mode-covering} nature of Forward KL forces the student to mimic this uncertainty, resulting in an \textit{entropy explosion} (orange curve) where the policy becomes overly diffused and loses precision. 
Conversely, Hard CE discards the teacher's soft probabilities entirely, forcing the student to rigidly track argmax targets. This causes a \textit{premature entropy collapse} (green curve), depriving the student of the action diversity necessary for effective state-space exploration and trapping it in local optima. 
In contrast, the \textbf{bounded mode-seeking} property of Reverse KL elegantly avoids both extremes. By filtering out the teacher's uncertain long tails while retaining sufficient stochasticity within the primary modes, Reverse KL maintains a healthy, stable entropy (blue curve). This ensures the agent remains decisive yet capable of exploration, translating to the highest and most stable task success rate.

\begin{figure}[htbp]
    \centering
    \includegraphics[width=0.55\textwidth]{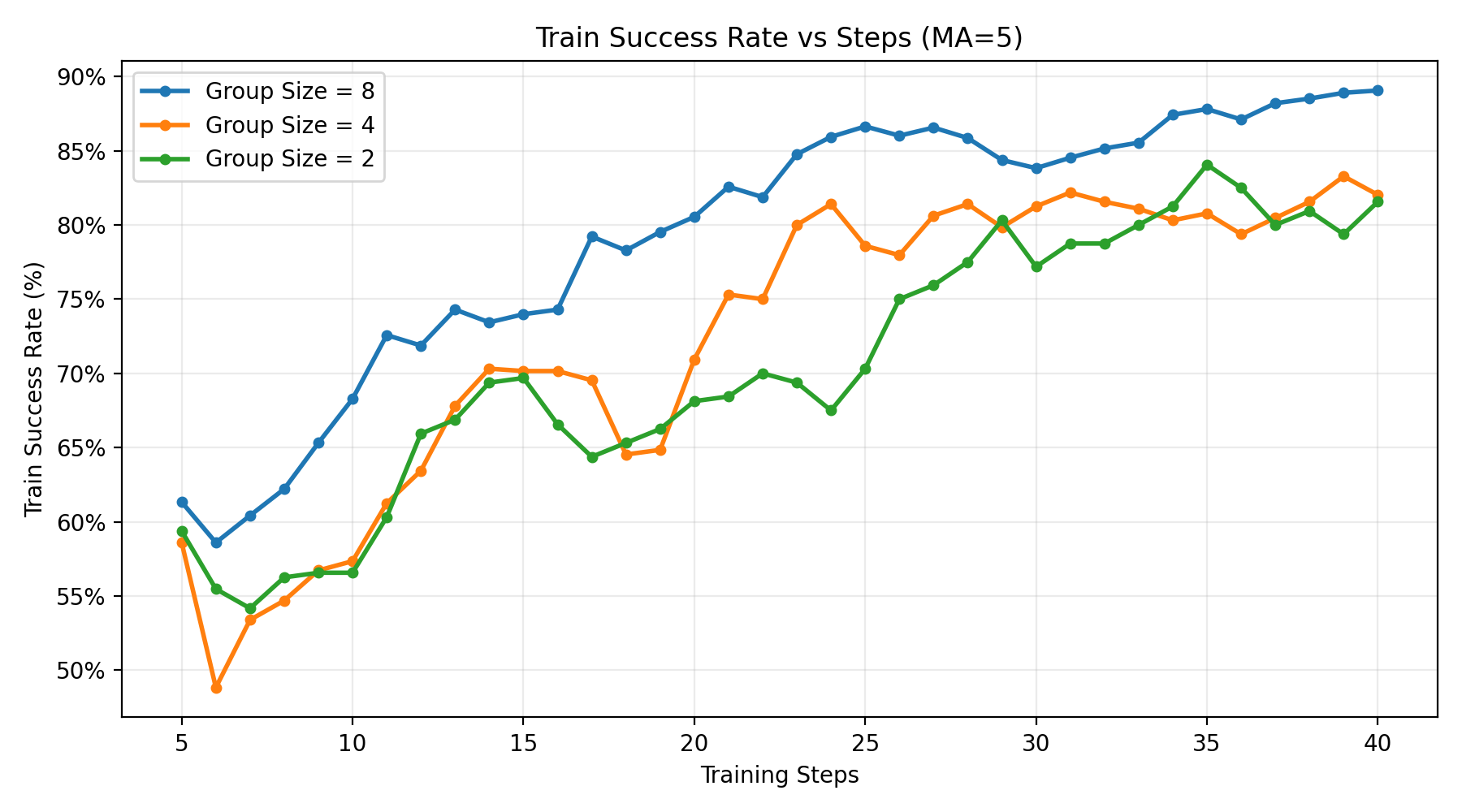}
    \caption{\textbf{Ablation on Group Sampling Size ($G$).} Training success rates demonstrate that while a larger group size ($G=8$) yields the smoothest optimization and highest final performance, smaller group sizes ($G=2, 4$) also achieve competitive success rates (over $80\%$) without performance collapse. This highlights a highly favorable trade-off between task performance and computational efficiency.}
    \label{fig:ablation_group_size}
    \vspace{-0.3cm}
\end{figure}

\noindent\textbf{2. Impact of Group Sampling Size ($G$).} 
We investigate the impact of the group-based sampling mechanism ($G$) on the LIBERO-Object suite with a fixed batch size of 32, evaluating $G \in \{2, 4, 8\}$. As illustrated in Figure~\ref{fig:ablation_group_size}, increasing the group size generally leads to smoother optimization. Specifically, $G=8$ achieves the highest final success rate ($\sim 89\%$). However, the most notable finding is that \textbf{smaller group sizes remain highly effective and do not lead to performance collapse}. Even with $G=2$, the success rate steadily climbs to over $80\%$, demonstrating competitive performance. This reveals a highly favorable trade-off between gradient variance and computational efficiency. Theoretically, a larger $G$ provides a more robust Monte Carlo approximation ($\mathbb{E}_{\tau \sim \pi_\theta}$) to average out environment stochasticity. Nevertheless, our results indicate that a minimal group size ($G=2$) still provides a sufficient signal-to-noise ratio. This offers a crucial practical advantage: \textbf{using smaller group sizes drastically reduces the computational overhead and wall-clock time} associated with environment rollouts and teacher inference.

%% file: sections/relatedwork.tex
\section{Related Work}
\label{sec:related_work}

\subsection{Offline SFT for Vision-Language-Action Models}
Recent advancements in robotic manipulation have been largely driven by VLA models~\cite{zitkovich2023rt,kim2024openvla,black2024pi0visionlanguageactionflowmodel,liu2024rdt,o2024open,song2025pd}, which are typically fine-tuned via offline SFT. While SFT benefits from dense supervision and fast convergence, it inherently suffers from covariate shift and compounding errors during online deployment. To mitigate this, interactive imitation learning algorithms like DAgger~\cite{kelly2019hg} collect expert annotations on student-induced out-of-distribution (OOD) states. However, adapting SFT to such on-policy settings typically relies on suboptimal alignment objectives. Standard DAgger uses hard expert labels (Hard-CE), which forces the student to rigidly track argmax targets, often leading to premature entropy collapse and poor exploration. Conversely, utilizing soft expert distributions implicitly optimizes the Forward-KL divergence. When the teacher exhibits high epistemic uncertainty in OOD states, Forward-KL forces the student to average across all modes, inevitably leading to entropy explosion and hesitant behaviors.

\subsection{Online RL for Vision-Language-Action Models}
To address the distribution shift inherent in offline SFT, Online RL has been introduced to align models with policy-induced state distributions~\cite{li2025simplevla,zang2025rlinf,lu2025vla,tan2025interactive,chen2025conrft,li2025vla,xu2025stare}. By continuously interacting with the environment, online RL exposes the policy to its own execution trajectories, naturally allowing the model to learn recovery behaviors when deviating from ideal paths, thereby enhancing closed-loop robustness. However, applying standard RL to billion-parameter VLAs is notoriously challenging due to the sparse nature of environment rewards, leading to prohibitively low sample efficiency and high-variance optimization.

Our work bridges this gap by framing VLA post-training as an on-policy RL problem augmented by dense teacher guidance. Unlike standard GRPO~\cite{li2025simplevla} that relies on sparse environmental signals, VLA-OPD utilizes a teacher model to provide token-level dense rewards. Crucially, inspired by recent insights in large language models~\cite{gu2023minillm,tan2023gkd}, we optimize a Reverse-KL objective for action prediction. This formulation naturally exhibits a mode-seeking property, encouraging the student policy to decisively commit to the teacher's most confident action mode in uncertain OOD states. Consequently, VLA-OPD achieves the robust exploration of RL without its extreme sample inefficiency, while gracefully avoiding the entropy extremes associated with the aforementioned imitation learning baselines.

%% file: sections/conclusion.tex
\section{Conclusion}
In this paper, we propose On-Policy VLA Distillation (VLA-OPD), a novel framework bridging the sample efficiency of offline SFT with the closed-loop robustness of online RL. By actively exploring the environment and leveraging token-level dense guidance from a teacher, VLA-OPD mitigates compounding errors and bypasses the extreme sample inefficiency of sparse-reward RL. A critical insight is identifying Reverse-KL as the optimal alignment objective. By avoiding the entropy extremes inherent to Forward-KL and Hard-CE in out-of-distribution states, Reverse-KL ensures stable and robust policy updates. This leads to superior success rates in complex robotic manipulation tasks without catastrophic forgetting. Future work will focus on reducing the framework's reliance on specific teacher models.